\pdfoutput=1

\documentclass[11pt]{article}

\usepackage[]{acl}

\usepackage{times}
\usepackage{latexsym}
\usepackage{url}

\usepackage[T1]{fontenc}

\usepackage[utf8]{inputenc}

\usepackage{microtype}
\usepackage{soul}
\usepackage{booktabs}
\usepackage{listings}
\usepackage[hang,flushmargin]{footmisc}
\newcommand{\bslabel}[1]{\smallskip\noindent\textbf{#1}}

\usepackage{graphicx}
\DeclareGraphicsExtensions{.pdf,.ai,.jpg,.png}
\setkeys{Gin}{pagebox=artbox}

\graphicspath{{./emnlp22-summary-workbench-figures/}}

\newsavebox\bscombox
\newcommand{\bscom}[3][]{%
  \sbox{\bscombox}{\fontsize{8}{9}\selectfont#1#2#3}
  \noindent
  \st{#2}{\fontsize{8}{9}\selectfont
    \color{blue}#3\ifx\\#1\\\else{\fontsize{8}{9}\selectfont\color{violet}[#1]}\fi
    }
  }

\begin{document}
\title{{\textsc{Summary Workbench}}\\ Unifying Application and Evaluation of Text Summarization Models}

\newcommand{\lei}{\textsuperscript{$\dagger$}}
\newcommand{\den}{\textsuperscript{$\ddagger$}}

\author{%
Shahbaz Syed 
\qquad Dominik Schwabe 
\qquad Martin Potthast  \\ [1.5ex]
\hspace{-5pt}Leipzig University\\
{\small\tt{<shahbaz.syed@uni-leipzig.de>}}}
\date{}

\maketitle
\begin{abstract}
This paper presents {\small\textsc{Summary Workbench}}, a new tool for developing and evaluating text summarization models. New models and evaluation measures can be easily integrated as Docker-based plugins, allowing to examine the quality of their summaries against any input and to evaluate them using various evaluation measures. Visual analyses combining multiple measures provide insights into the models' strengths and weaknesses. The tool is hosted at \url{https://tldr.demo.webis.de} and also supports local deployment for private resources.
\end{abstract}

\section{Introduction}

Automatic text summarization reduces a long text to its most important parts and generates a summary. 
Usually, a learning-based summarization model is developed in two basic steps: \emph{model development} and \emph{model evaluation}. Given a collection of documents accompanied by one or more human-written (reference) summaries, first a set of features representing the documents is manually created or automatically extracted through supervised learning. The resulting model is then used to generate one or more (candidate) summaries, which are analyzed manually and/or with evaluation measures for their similarity to the reference summaries. These steps are iterated, optimizing the model and its parameters using a validation set. The models that perform best in the validation are selected for evaluation on the test set. With standardized test sets for each document collection, comparisons with models created earlier are reported.

However, these steps are associated with comparatively tedious tasks: During model development, summaries of individual documents are often generated and immediately evaluated to identify deficiencies and improve the model, including comparisons to other models. The latter requires third-party models to be operational despite their heterogeneous software stacks. Such ``on-the-fly evaluation'' during development entails that candidate and reference summaries as well as source documents are analyzed manually or by automatic measures. This multi-text comparison is often not supported by visualization, although this leads to a better understanding of the content coverage and possible selection biases of a model~\cite{vig:2021,syed:2021}. The analysis of evaluation results for model selection also benefits from visual support~\cite{tenney:2020}. Previous research in the field of automatic summarization has not yet resulted in a unified set of tools for these purposes which is the main goal of this paper. 

With {\small\textsc{Summary Workbench}}, we introduce the first unified combination of application and visual evaluation environments for text summarization models. Currently, it integrates 15~well-known summarization models (26~variants in total) and 10~standard evaluation measures from the literature. With \textbf{FeatureSum}, it also includes a new feature-based extractive summarization model that implements features from the literature predating the deep learning era. Underlying all of the above is a specification and interface that allows easy integration of new models and measures to facilitate large-scale experiments and their reproducibility.

In what follows, Section~\ref{related-work} reviews related work on tools to assist summarization research and development. Section~\ref{design} overviews the key design principles of the {\small\textsc{Summary Workbench}}, and Section~\ref{models-and-measures} provides a complete overview of all the models and measures hosted to date. Included are general-purpose models, guided models that accept user prompts to guide summary generation, and models tailored to argumentative language and to news articles. A wide range of commonly employed evaluation measures are included, covering both lexical as well as semantic overlap measures.%
\footnote{
Source code is available at \url{https://github.com/webis-de/summary-workbench}.
}


\section{Related Work}
\label{related-work}

The development of tools for summarization research has gained momentum recently, and several tools have been presented for (sub)tasks of the two steps above: Tools such as HuggingFace~\cite{wolf:2020}, FairSeq~\cite{ott:2019}, SummerTime~\cite{ni:2021}, TorchMetrics~\cite{detlefsen:2022}, SacreROUGE\cite{deutsch:2020}, PyTorch Hub,%
\footnote{\url{https://pytorch.org/hub/}}
and TensorFlow Hub%
\footnote{\url{https://www.tensorflow.org/hub/}}
focus on hosting several state-of-the-art text summarization models and automatic evaluation measures. These tools have significantly improved accessibility to working models. However, only some provide a very minimal interface for inference of summaries and their online/offline comparative analyses. Many authors also choose to share their models independently, be they on GitHub or elsewhere, as standalone repositories instead of integrating with any tools. To lower the bar of (latter) integration as much as possible, {\small\textsc{Summary Workbench}} simplifies model and measure integration as plugins (using Docker). In this way, models under development or private ones can be locally compared to others and can be archived together with all their dependencies for reproducibility. Similar efforts have been made in the information retrieval community via the Docker-based toolkit such as Anserini~\cite{yang:2018}. 


Tools such as LIT~\cite{tenney:2020}, Summ\-Vis~\cite{vig:2021}, and Summary Explorer~\cite{syed:2021} focus on qualitative model evaluation by providing static visual analyses of the relation between the summary and its source document. {\small\textsc{Summary Workbench}} adapts some of their visualizations next to new ones, and complements them with interactive visual analytics for quantitative evaluation according to multiple measures. Users can explore the distribution of scores, select data points of interest and inspect them in relation to the source document to better understand the dataset. 

The success of past summarization research and development has relied a lot on in-depth manual error analyses. This being one of the most laborious tasks in every natural language generation evaluation, we believe that visually comparing summaries from multiple models for many different texts, and contextualizing manual review with  multiple measures is crucial to both scale up error analysis, and to better understand the capabilities and limitations of the technology. As this still requires juggling many different, incompatible tools, the unified approach of {\small\textsc{Summary Workbench}} aims at lowering the bar for scaling up interactive experimentation.

\section{An Interactive Visual Summarization Model Development \& Evaluation Tool}
\label{design}

{\small\textsc{Summary Workbench}} implements two interactive views corresponding to the two basic summarization model development steps: a \emph{summarization view} and an \emph{evaluation view}.

\begin{figure*}[ht]
\includegraphics[width=\textwidth]{summarize-view-titles-above}\vspace*{-4ex}
\caption{Two key components of the summarization view: On the left, an input text can be summarized via multiple extractive and abstractive summarization models; lexical overlap is highlighted on demand for each candidate summary and can be adjusted for varying n-gram lengths. On the right, content agreement among summaries from different models; any summary can be selected as the reference against which the others can be visually compared.}
\label{summarize-view-titles-above}
\end{figure*}

\subsection{Summarization View}

Figure~\ref{summarize-view-titles-above} shows the summarization view, where multiple extractive/abstractive summarization models can be used to summarize texts, web pages, or scientific documents on demand, controlling for summary length. For scientific documents, relevant sections from a given paper to be summarized can be chosen. Explicit guidance signals for focused summarization can be provided as input to corresponding models (reviewed in Section~\ref{models-and-measures}). 

Generated summaries (candidates) can be visually inspected for their lexical overlaps (highlighted on demand) with their source document or with another summary. This provides a quick overview of the models' effectiveness at capturing important content as well as any factual errors prevalent in abstractive summarization~\cite{maynez:2020}. Additional functionalities include uploading multiple documents to be summarized via a single file, and command line access to all models.

\begin{figure*}[ht]
\includegraphics[width=\textwidth]{evaluate-view-titles-above}\vspace*{-2ex}
\vspace{-15ex}
\caption{Two key components of the evaluation view: On the left, a text overlap viewer displays content coverage of the summaries in relation to the source document via lexical and semantic overlap (via Spacy embeddings). On the right, an interactive plotter allows selecting examples with specific scores for a combination of evaluation measures. Additionally, the distribution of scores is also shown.}
\label{evaluate-view-titles-above}
\end{figure*}

\subsection{Evaluation View}
\label{evaluate-view}

Figure~\ref{evaluate-view-titles-above} shows the evaluation view, where candidate summaries are compared with reference summaries using multiple lexical/semantic content overlap measures. Candidate summaries from multiple models, either generated using the summarization view or uploaded as a file where each example is encoded as {\small{$\tt<doc, ref, c_1, c_2, ..., c_n>$}} can be evaluated. Lexical/semantic overlap of candidate/reference summaries {\small $\tt c_i$}/{\small $\tt ref$} with the source document {\small $\tt doc$} can also be visualized. Computed scores can be neatly exported as~CSV or \LaTeX~tables.

Scores from the chosen evaluation measures can be further explored through an \emph{interactive plotter}. Among other things, the plotter allows users to visually correlate different evaluation measures, identify outliers/challenging source documents or strongly abstractive summaries among the candidates. This facilitates a deeper understanding of the quantitative performance of the models as well as an understanding of the evaluation datasets. Two more use cases of the interactive plotter are explained in Section~\ref{use-cases}.

\subsection{Plugin Server}
\label{plugin-server}

New summarization models and evaluation measures are integrated as container-based plugins. A model/measure plugin can either be a local directory or a remote Git~repository containing specification of dependent software and data (checkpoints, embeddings, lexicons), and implementations of the interfaces SummarizerPlugin and MeasurePlugin. Model metadata such as name, type, version, source, citation, and other custom arguments are provided as YAML~configurations. Each plugin runs inside a Docker container with its own server that handles API~calls following the OpenAPI~specification.%
\footnote{\url{https://www.openapis.org}}
This setup allows users to safely self-host the entire application. Developed plugins can be easily shared with the community via DockerHub images or Git~repositories. Examples are found in our tool's technical documentation.%
\footnote{\url{https://webis.de/summary-workbench/}}

\begin{figure*}
\includegraphics[width=\textwidth, clip]{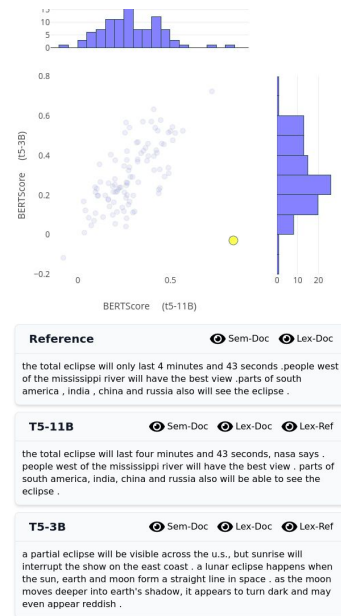}
\caption{Two example use cases of interactive plotter of the evaluation view: On the left, correlations between pairs of evaluation measures are analyzed. On the right, abstractive summaries from two variants of the T5~model for the chosen data point (highlighted yellow) are shown.}
\label{figure-use-cases}
\end{figure*}

\section{Models and Measures}
\label{models-and-measures}

{\small\textsc{Summary Workbench}} hosts 15~extractive/abstractive summarization models and 10~lexical/semantic evaluation measures for English text. Each of these is configured as a Docker-based plugin that can be customized and instantiated accordingly. For details on model checkpoints, see Appendix~\ref{model-details}.

\subsection{Summarization Models}

We provide a diverse set of models applicable to multiple text domains such as news, argumentative texts, web pages, and product reviews. Model types include extractive, abstractive, supervised, unsupervised, and guided summarization, the latter requiring additional user input. 

\subsection*{General-purpose Summarization}

Models that work in an unsupervised fashion or leverage external knowledge via contextual embeddings are supposed to be capable of summarizing any kind of text. We provide the following models suitable for general-purpose text summarization.

\bslabel{FeatureSum} is our new extractive summarization model which scores a sentence in the text based on a combination of standard features to identify key sentences~\cite{luhn:1958,nenkova:2012}: TF-IDF, content units (named entities, noun phrases, numbers), position in text, mean lexical connectivity (number of tokens shared with the remaining sentences), ratio of words that are not stop words, length (relative to the longest sentence in the text), and word overlap with the title. The final score of a sentence is the product of the individual feature values. Sentences are then ranked based on these scores to produce the final summary. Different combinations of these features can be chosen by simply toggling them in the interface. This also allows for dynamically reproducing existing models from the literature provided their specific feature sets are available.

\bslabel{TextRank}~\cite{mihalcea:2004} is a graph-based model which employs PageRank~\cite{brin:1998} on the document graph consisting of sentences as nodes to compute the strength of their connections. Top-ranked sentences within a length budget are taken as the extractive summary. We also provide the two variants \textbf{PositionRank} and \textbf{TopicRank}, which consider the sentence position and its overlap with topic sentence (document's title or its first sentence) to compute the ranking via PyTextRank~\cite{nathan:2016}.

\bslabel{BERTSum}~\cite{miller:2019} employs contextual embeddings from BERT~\cite{devlin:2019} to extract key sentences in an unsupervised fashion by first clustering all sentence embeddings using k-means~\cite{hartigan:1979} and then retrieving those closest to the centroids as the summary.

\bslabel{PMISum}~\cite{padmakumar:2021} is an unsupervised extractive model that includes measures to score the relevance and redundancy of the sentences of the source document. These measures are based on pointwise mutual information~(PMI) computed by pre-trained language models. Summary sentences are selected via a greedy algorithm to maximize relevance and minimize redundancy.

\bslabel{LoBART}~\cite{manakul:2021} addresses the input length limitations of transformers~\cite{vaswani:2017} that restrict capturing long-span dependencies in  long document summarization. Local self-attention and explicit content selection modules are introduced to  effectively summarize long documents such as podcast transcripts and scientific documents.

\bslabel{Longformer2Roberta} effectively combines Longformer~\cite{beltagy:2020}, developed for processing long documents, and RoBERTa~\cite{liu:2019a}, a robustly trained BERT model as the decoder, based on leveraging pre-trained checkpoints of large language models~\cite{rothe:2020}.

\subsection*{Guided Summarization}

The following models accept explicit inputs provided by users to guide the summarization process towards generating user-specific summaries.

\bslabel{Biased TextRank}~\cite{kazemi:2020} is an extension of the TextRank model which takes an explicit user input as the ``focus'', represented via contextual embeddings to guide the ranking of the document sentences. Summary extraction is based on the semantic alignment between the document sentences and the provided focus signal.

\bslabel{GSum}~\cite{dou:2021} is a guidance-based abstractive model that takes different types of external guidance signals: text inputs, highlighted sentences, keywords, or extractive oracle summaries derived from the training data. These signals along with the source text are used to generate focused and faithful abstractive summaries.

\subsection*{Argument Summarization}

Summarizing argumentative texts (opinions, product reviews) requires that the model be able to identify high-quality, informative, and argumentative sentences from the text. We provide three models specifically developed for this task.

\bslabel{ArgsRank}~\cite{alshomary:2020} is an extractive model for creating argument snippets. It augments TextRank with two new criteria: \emph{centrality in context} and \emph{argumentativeness} to help the model retrieve important and argumentative sentences.

\bslabel{ConcluGen}~\cite{syed:2021a} is a transformer model for generating informative conclusions of argumentative texts by balancing the trade-off between abstractiveness and informativeness of the output. It was finetuned on the Conclugen corpus comprised of pairs of argumentative text and a human-written conclusion.

\bslabel{COOP}~\cite{iso:2021} is an unsupervised opinion summarization model that employs latent vector aggregation by searching for optimal input combinations of sentence embeddings to address the summary vector degeneration problem caused by simple averaging. Specifically, it finds convex combinations that maximize the word overlap between the source document and its summary.

\subsection*{News Summarization}

A majority of the existing summarization models are trained on news datasets, since news have been and are readily available. These models have shown strong performance in creating fluent abstractive summaries~\cite{huang:2020}. We provide the following models for summarizing news.

\bslabel{BART}~\cite{lewis:2020} is a transformer denoising autoencoder for pre-training sequence-to-sequence models. Its main objective is to reconstruct the source text corrupted by employing arbitrary noising functions (masking text spans, randomly shuffling sentences) which helps the model learn better representations of the source texts for text summarization~\cite{huang:2020}.

\bslabel{T5}~\cite{raffel:2020} is a unified text-to-text transformer model that exploits the strengths of transfer learning on a variety of problems that can be modeled as text generation tasks. A task-specific prefix is added to each input sequence (e.g.,~``summarize:<document>'') that teaches the model to summarize accordingly.

\bslabel{Pegasus}~\cite{zhang:2020} is a transformer model pre-trained with a self-supervised summari\-zation-specific training objective called ``gap-sen\-tences generation'': important sentences are removed/masked from the source text and must be jointly generated as output from the remaining sentences, similar to an extractive summary.

\bslabel{CLIFF}~\cite{cao:2021} leverages contrastive learning for generating abstractive summaries that are faithfully and factually consistent with the source texts. Reference summaries are used as positive examples while automatically generated erroneous summaries are used as the negative examples for training the model.

\bslabel{Newspaper3k} is an open-source library for extracting news articles from the web which provides a module for extractive summarization that ranks sentences based on keywords and title words.%
\footnote{\url{https://newspaper.readthedocs.io/en/latest/}} 

\subsection{Evaluation Measures}

Evaluation measures for summarization typically quantify the lexical/semantic overlap of a candidate summary with a reference summary. We provide the following measures covering both. 

\subsection*{Lexical Measures}

Measures based on lexical overlap return precision, recall, or F1~scores on varying granularities of text between the candidate summary and one or more reference summaries.

\bslabel{BLEU}~\cite{papineni:2002} is a standard measure for machine translation adapted for summarization. It includes a brevity penalty to account for length differences while computing n-gram overlap.

\bslabel{ROUGE}~\cite{lin:2004} is the most common measure for summarization which computes precision, recall, and F1~scores based on n-gram overlap, where n-grams include unigrams, bigrams, and the longest common subsequence.

\bslabel{METEOR}~\cite{banerjee:2005} aligns a candidate with a set of references by mapping each unigram of a candidate to 0/1~unigrams of the reference based on exact, stem, synonym, and paraphrase matches. It then computes precision, recall, and F9~scores (i.e., weighted harmonic mean, strongly emphasizing recall) based on that.

\bslabel{CIDEr}~\cite{vedantam:2015} is a consensus-based measure (originally for evaluating image captioning) which measures the similarity of a candidate against a set of references by counting the frequency of the common n-grams of a candidate.

\subsection*{Semantic Measures}

Measures based on semantic overlap compute the semantic alignment between candidates and references at the token/word/sentence level based on their static/contextual embeddings.

\bslabel{Greedy Matching}~\cite{rus:2012} aligns a candidate and a reference by greedily matching each candidate word to a reference word based on their embeddings' cosine similarity. Average similarity over all candidate words aligned to reference words and vice versa are computed whose average is the final score.

\bslabel{MoverScore}~\cite{zhao:2019} combines contextual embeddings from BERT using the word mover's distance~\cite{kusner:2015} to compare a candidate against a set of references by considering both the amount of shared content as well as the extent of deviation between them.

\bslabel{BERTScore}~\cite{zhang:2020a} computes a similarity score for each candidate token with each reference token using contextual embeddings from~BERT. The measure is also robust to adversarial modifications of the generated text.

\bslabel{BLEURT}~\cite{sellam:2020} is a learned measure based on BERT that models human judgments with a few thousand biased training examples. The model is pre-trained using millions of synthetic examples created via scores from existing measures (BLEU, ROUGE, BERTScore), and textual entailment, for better generalization.

\bslabel{BARTScore}~\cite{yuan:2021} uses the weighted log probability of generating one text given another to compute faithfulness (source~$\rightarrow$~candidate), precision (reference~$\rightarrow$~candidate), recall (candidate~$\rightarrow$~reference), and the F1~score.

\bslabel{CosineSim} includes two embedding-based cosine similarity measures using Spacy word vectors~\cite{honnibal:2020} and Sentence-BERT~\cite{reimers:2019}.

\begin{figure*}
\includegraphics[width=\textwidth]{color-schemes}
\vspace*{-4ex}
\caption{Customization options available for visualization of document and summary overlap. Users can select the minimum word overlap, preserve duplicate words, and ignore stop words to be visualized. Also, they can instantly preview each color scheme and set it as their default. The tool provides colorful, soft gradient-based, and grayscale schemes to account for color blindness.}
\label{figure-color-schemes}
\end{figure*}
    
\section{Interaction Use Cases}
\label{use-cases}

Figure~\ref{figure-use-cases} shows two use cases of the interactive plotter. First, users can analyze any correlation between two measures of choice for a summarization model. Here, we find that MoverScore and BERTScore have strong correlation as they both employ contextual embeddings from BERT to compute the overlap between candidate and reference summaries. Likewise, we find that the static token embeddings from Spacy have a broader distribution of scores in comparison. 

As a second use case, the interactive plotter allows comparing two variants of the same model architecture using any measure. Here, we inspect the T5~model (its~3B and 11B~variants) using BERTScore to find that the larger variant generates a summary very similar to the reference while the smaller variant creates a summary that is topically related but not accurate in comparison to the reference.

\section{Conclusion}

In this paper we present {\small\textsc{Summary Workbench}}, a tool that unifies the application and evaluation of text summarization models. The tool supports integrating summarization models and evaluation measures of all kinds via a Docker-based plugin system that can also be locally deployed. This allows safe inspection and comparison of models on existing benchmarks and easy sharing with the research community in a software stack-agnostic manner. We have curated an initial set of 15~models (26~including all variants) and 10~evaluation measures and welcome contributions from the text summarization community. An extension of the tool's features to related text generation tasks such as paraphrasing and question answering is foreseen. 

\section{Ethical Statement and Limitations}

Our tool builds on open source models and evaluation measures contributed by the corresponding authors. We expect all users of our tool to diligently cite the authors of all models and measures when they use them via our tool, instead of just citing our tool. The tool provides direct links to the relevant sources for each hosted summarization model and evaluation measure to facilitate this.

The models and measures may have intrinsic biases, which ideally, our tool may help to identify. However, our tool itself may have biases, especially with respect to its visualizations: Visualization in general is a difficult task, and visual analytics for data analysis in particular may lead to invalid conclusions if the underlying visualization itself is flawed. Although we did our best to avoid any non-standard visualizations and relied on widely used tools to plot them, we caution users of possible errors from either the dependent libraries or their integration in our tool. Validating a visual analytics tool poses non-trivial research tasks of its own right, which we leave for future work. We do hope that the community will diligently report any errors they may encounter.

To account for color blindness, we strived to provide multiple (soft) gradient-based color schemes and grayscale colors (Figure~\ref{figure-color-schemes}) for all our visualizations. Instant previews of each color scheme are available to help users customize the tool's visuals.

\bibliography{emnlp22-summary-workbench-lit}
\bibliographystyle{acl_natbib}
\appendix
\section{Model Details}
\label{model-details}
\begin{table}[th!]
\centering
\small
\renewcommand{\arraystretch}{1.2}
\begin{tabular}{@{}lp{4cm}@{}}
\toprule
\bfseries{Summarizer}         & \bfseries{Model}                             \\ 
\midrule
BERTSummarizer     & distilbert-base-uncased                                 \\
LoBART             & podcast\_4K\_ORC                                        \\
Longformer2Roberta & patrickvonplaten/longformer2roberta-cnn\_dailymail-fp16 \\
ConcluGen          & dbart                                                   \\
CLIFF              & pegasus\_cnndm                                          \\
COOP               & megagonlabs/bimeanvae                                   \\
BART               & facebook/bart-large                                     \\
Pegasus            & google/pegasus                                          \\
T5-Base            & huggingface.co/t5-base                                  \\
\midrule
\bfseries{Evaluator}          & \bfseries{Model}                             \\
\midrule
BARTScore          & facebook/bart-large-cnn                                 \\
Spacy Similarity   & en\_core\_web\_lg                                       \\
SBERT              & roberta-large-nli-stsb-mean-tokens                      \\
BLEURT             & bleurt-base-128                                         \\
BERTScore          & roberta-large-mnli                                      \\
Greedy Matching    & glove.6B.300d                                           \\
MoverScore         & MoverScoreV1                                            \\ 
\bottomrule
\end{tabular}
\end{table}
\end{document}